\documentclass[review]{elsarticle}

\usepackage{lineno,hyperref}
\modulolinenumbers[5]

\journal{Journal of \LaTeX\ Templates}

\begin{document}

\begin{frontmatter}

\title{On the role of words in the network structure of texts: application to authorship attribution}

%%%%%%%%%%%%%%%%%%%%%%%%%%%%%%%%%%%%%%%%%%%%%%%%%%%%%%%%%%%%%%%%%%%%%%%%%%%%%%%%%%%%%%%%%%%%%%%%%%%%%%%%%%%%%%%%%%%%%%% BEGIN

\author[ifscaddress]{Camilo Akimushkin\corref{mycorrespondingauthor}}
\cortext[mycorrespondingauthor]{Corresponding author}
%\ead{camilo.akimushkin@gmail.com}

\author[icmcaddress]{Diego R. Amancio}
\ead{diego.raphael@gmail.com}

\author[ifscaddress]{Osvaldo N. Oliveira Jr.}
%\ead{chu@ifsc.usp.br}

\address[ifscaddress]{S\~ao Carlos Institute of Physics, University of S\~ao Paulo, Avenida Trabalhador S\~ao-carlense 400, S\~ao Carlos, S\~ao Paulo, Brazil}
\address[icmcaddress]{Institute of Mathematics and Computer Science, University of S\~ao Paulo, Avenida Trabalhador S\~ao-carlense 400, S\~ao Carlos, S\~ao Paulo, Brazil}
%\address[uspaddress]{Avenida Trabalhador São-carlense 400, São Carlos, São Paulo, Brazil}

%%%%%%%%%%%%%%%%%%%%%%%%%%%%%%%%%%%%%%%%%%%%%%%%%%%%%%%%%%%%%%%%%%%%%%%%%%%%%%%%%%%%%%%%%%%%%%%%%%%%%%%%%%%%%%%%%%%%%%% END

\begin{abstract}
Well-established automatic analyses of texts mainly consider frequencies of linguistic units, e.g. letters, words and bigrams, while methods based on co-occurrence networks consider the structure of texts regardless of the nodes label (i.e. the words semantics). In this paper, we reconcile these distinct viewpoints by introducing a generalized similarity measure to compare texts which accounts for both the network structure of texts and the role of individual words in the networks. We use the similarity measure for authorship attribution of three collections of books, each composed of $8$ authors and $10$ books per author. High accuracy rates were obtained with typical values from $90\%$ to $98.75\%$, much higher than with the traditional the TF-IDF approach for the same collections. These accuracies are also higher than taking only the topology of networks into account. We conclude that the different properties of specific words on the macroscopic scale structure of a whole text are as relevant as their frequency of appearance; conversely, considering the identity of nodes brings further knowledge about a piece of text represented as a network.
\end{abstract}


\end{frontmatter}

%\linenumbers

\section{Introduction}

The huge volume of written text produced everyday makes it imperative to use automatic tools to retrieve relevant information, e.g. with text summarization, information retrieval methods, polarity analysis, citation analysis, and document classification~\cite{Liang2017802,0295-5075-98-5-58001,Zhong2017462,manning2008introduction,10.1371/journal.pone.0171649,Inform2012427,Viana2013371}. An essential step in many of these tasks is to compare pieces of texts, as in classification of texts into categories~\cite{manning2008introduction} and in search engines where typically a list of texts relevant to a given query is retrieved.  A special case is the pairwise comparison, where one searches for similarities between pairs of texts, which is actually a typical subtask in the authorship attribution process~\cite{juola2006authorship}. Automatic authorship attribution has been made with varied strategies~\cite{Stamatatos:2009:SMA:1527090.1527102}, from the use of first-order statistics of linguistic elements to the processing of text represented as networks~\cite{amancio2011comparing,1742-5468-2015-3-P03005}. For example, the frequency of characters~\cite{peng2003language,escalante2011local}, phonemes~\cite{forstall2010features}, and morphemes~\cite{kukushkina2001using,chaski2005s} has been explored, with texts normally modelled as lists of individual words, i.e. word order is disregarded. The archetype of such models is the so-called bag-of-words (BoW) model~\cite{Harris:54}, where the text is represented as the set of its constitutive words by counting the number of appearances for each word. Word frequencies, which follow Zipf's law~\cite{zipf1935psycho,ferrer2001two}, can then be used straightforwardly as attributes in a machine learning scheme~\cite{amancio2014systematic} or to further build specific similarity measures.

Variations of the BoW model have been developed to address possible biases, e.g. the tendency of larger texts of being more likely to be considered similar to any other. These variations include the use of the term frequency-inverse document frequency (TF-IDF) statistic~\cite{sparck1972statistical,manning2008introduction}, where lower relevance is assigned to words frequent in the document as well as in the whole collection. The model has also been modified to incorporate other kinds of data, such as in the bag-of-features model used for image analysis~\cite{zhang2007local}. Another important modification is to consider $n$-grams, i.e., groups of $n$ adjacent words ~\cite{kevselj2003n,clement2003ngram}, in an attempt to take syntactic information into account, since the BoW model disregards word ordering. In other types of work, the syntactic roles of the words in sentences are used for authorship attribution~\cite{baayen1996outside,rygl2012authorship}. It must be noted, nevertheless, that all of these approaches are based on the counting of features, even if some consider small-scale structural relationships.

An alternative perspective has been developed in recent years from the discovery that language features may be best described by complex networks models~\cite{Cong2014598}. The structure of a text, for instance, can be mapped onto a co-occurrence network~\cite{amancio2011comparing}, which is characterized by power-law distributions \cite{ferrer2001two,dorogovtsev2001language}, and core-periphery structures~\cite{choudhury2010global}. Even though the general features of these complex networks remain analogous for texts in the same language, the network representation can also be used for classification tasks, particularly for authorship attribution~\cite{amancio2011comparing,mehri2012complex,akimushkin2017plos}.

While the frequency-based methods overlook all structural relationships among words farther than in the same sentence, the methods based on co-occurrence networks ignore the identity of the words (i.e. which actual word corresponds to a given node), thus characterizing the texts only on the basis of the network topology. In this study, we reconcile both viewpoints to show that, from a network perspective, words can play relevant roles in the structure of a text besides their frequencies.

\section{Methods}

The methodology proposed to address the authorship attribution task consists of four steps: i) construct a co-occurrence network for each text; ii) obtain various distance matrices for the collection using the proposed similarity metrics (see below); iii) join the various distance matrices with multi-dimensional scaling~\cite{BorgGroenen2005}; and iv) analyze the resulting data with standard supervised learning algorithms~\cite{amancio2014systematic}. These steps are described in detail below. The model was applied to three collections of $80$ literary texts. Each collection contains $10$ texts per author for $8$ authors from the 19th century, with $22$ of the $24$ authors being native English writers (details of the collections are included in the Supporting Information).

\subsection{Network construction and characterization}

Texts are pre-processed for constructing the networks, with stopwords, such as articles and prepositions, being removed, and lemmatization being applied to reduce different forms to a common base form. Lemmatization is assisted by a part-of-speech tagger based on entropy maximization~\cite{GreRub}, in order to solve ambiguities in mapping words to their lemmatized form. From the resulting pre-processed text, a co-occurrence (or word adjacency) network is built, where each distinct word is a node and two nodes are connected if the words appear consecutively in the text. The link is directed according to the natural reading order. For instance, the title of this paper generates the network: role $\rightarrow$ word $\rightarrow$ network $\rightarrow$ structure $\rightarrow$ text $\rightarrow$ application $\rightarrow$ author $\rightarrow$ attribution. Each link has a default weight equal to one, which is increased by one unit each time the pair of words appears in the text.

Networks were characterized in this study by four well-known node-local metrics:

\begin{enumerate}

\item Degree ($k_i$):  this metric corresponds to the number of links attached to a node. As a consequence of the construction rules imposed by co-occurrence networks, there is strong correlation between this metric and the word frequency.

\item Average shortest path length ($l_i$): this is the typical distance between two nodes of the network, given by:
\begin{equation}
l_i = N^{-1} \Sigma_j d_{ij},
\end{equation}
where $d_{ij}$ is the shortest path length between nodes $i$ and $j$, and $N$ is the number of nodes. This metric is useful to identify keywords in written texts, irrespectively of the word frequency~\cite{0295-5075-57-5-759}. Low values of $l$ are not only associated to the frequent words, but also to the words appearing close to other relevant words in the text.

\item Betweenness centrality ($B_i$): the betweenness is the fraction of all shortest paths that pass through the node, i.e.
\begin{equation}
B_i = \sum_{i\neq j\neq k} \frac{n_{jk}^{(i)}}{n_{jk}},
\end{equation}
where $n_{jk}^{(i)}$ is the number of shortest paths from $j$ to $k$ passing through $i$ and $n_{jk}$ is the total number of shortest paths from $j$ to $k$. In text analysis, the betweenness can be interpreted as a measure to quantify the ability of a word to appear in restrict or wider contexts~\cite{amancio2011comparing}.

\item Intermittency ($I_i$): the intermittency is a measure that quantifies the spatial distribution of a given word along a text. To define this measure, consider the text as a sequence of tokens. This sequence generates, for each word $i$, a time series $T^{(i)} = \{t^{(i)}_1, t^{(i)}_2,\ldots,t^{(i)}_{f_i}\}$, where $t^{(i)}_j$ corresponds to the position of the $j$-th occurrence of the word $i$. The interval recurrence ($\tau$) for word $i$ is defined as the spatial difference between two occurrences, i.e. $\tau^{(i)}_j = t^{(i)}_j - t^{(i)}_{j-1}$. The set of all values of $\tau^{(i)}_j$, i.e. $\mathcal{T}^{(i)} = \{\tau^{(i)}_1, \tau^{(i)}_2,\ldots\}$ is used to quantify the regularity of the appearance of $i$ along the sequence of tokens. More specifically, this regularity is computed using the intermittency defined as:
\begin{equation}
I_i = \sigma_\mathcal{T}/\langle \mathcal{T} \rangle = \Bigg{[} \frac{\langle \mathcal{T}^2 \rangle}{\langle \mathcal{T} \rangle^2} - 1 \Bigg{]}^{1/2},
\end{equation}
where $\sigma_\mathcal{T}$ and $\langle \mathcal{T} \rangle$ are the standard deviation and average of $\mathcal{T}$, respectively. In text networks, the intermittency also measures the relevance of words, since it has been shown that intermittent (i.e. bursty) words are the ones most related to the subject being approached~\cite{0295-5075-57-5-759}.
%is the interval between successive occurrences of the word in the text and is associated to the lengths of closed loops in the network. {Note que naoda pra entender o que eh Delta. Nao foi dito que ÃƒÂ© uma sÃƒÂ©rie temporal de intervalos de recorrencia}.

\end{enumerate}

\subsection{Similarity metrics}

The novelty introduced in this work is to compare the words representing the most relevant nodes in the network topology, in contrast to previous approaches where only the statistics of topological metrics were taken into account~\cite{amancio2011comparing}. We consider as the most relevant the nodes possessing the highest degree and betweenness. As for the other metrics, namely average shortest paths and intermittency, we chose the nodes with lowest values. We tested the largest shortest paths but the results were not as good. Intermittency obeys a power law with positive exponent. We hypothesize that two pieces of text will be similar if there is significant overlap in the words (nodes) considered most relevant in both texts.

\begin{figure}[t]
\centering
\includegraphics[width=0.75\columnwidth]{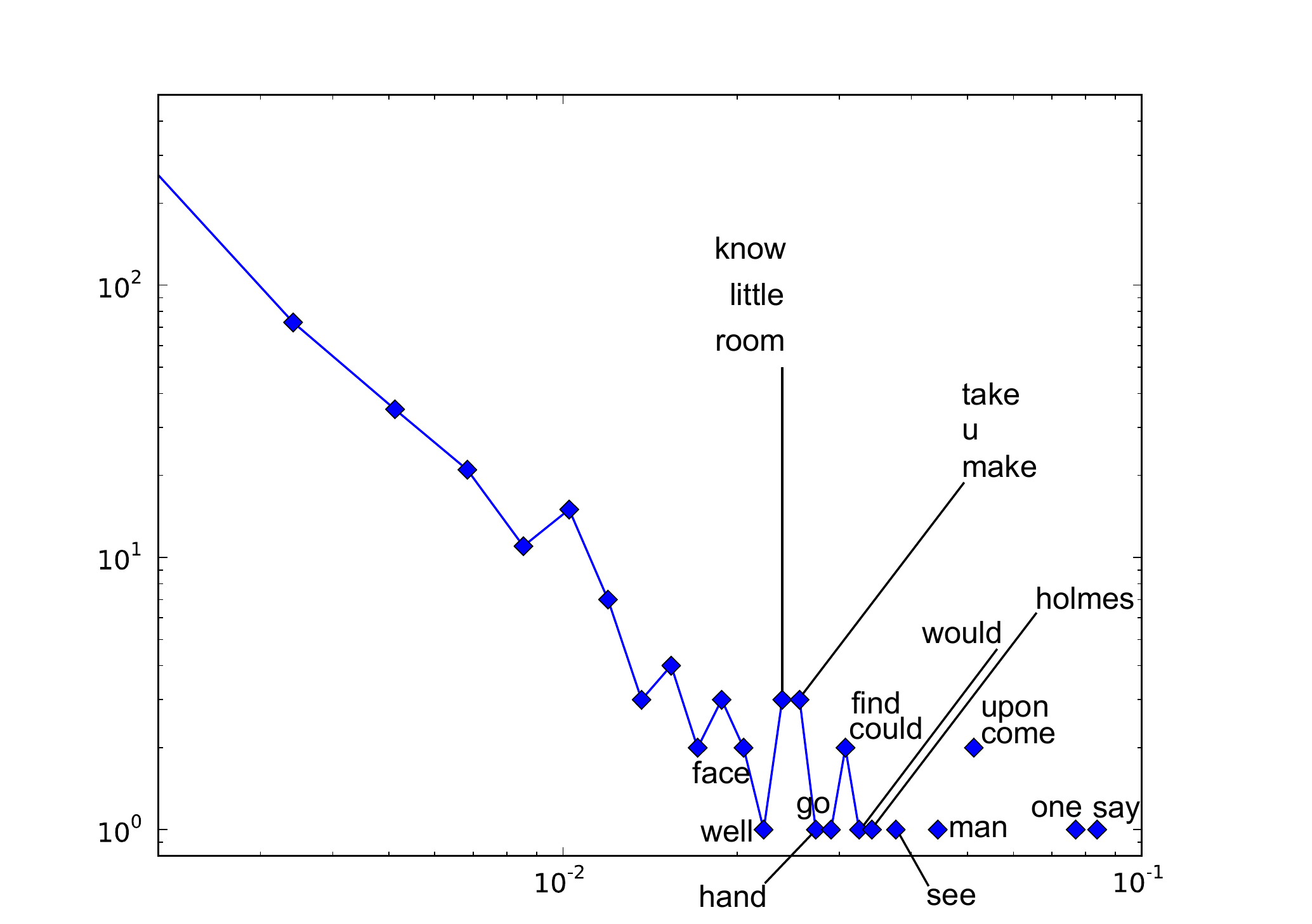}
\caption{Betweenness centrality distribution for ``The Memoirs of Sherlock Holmes''. The top 20 nodes are labeled by the corresponding words.}
\label{fig:bc_distibution_1}
\end{figure}

\begin{figure}[t]
\centering
\includegraphics[width=0.75\columnwidth]{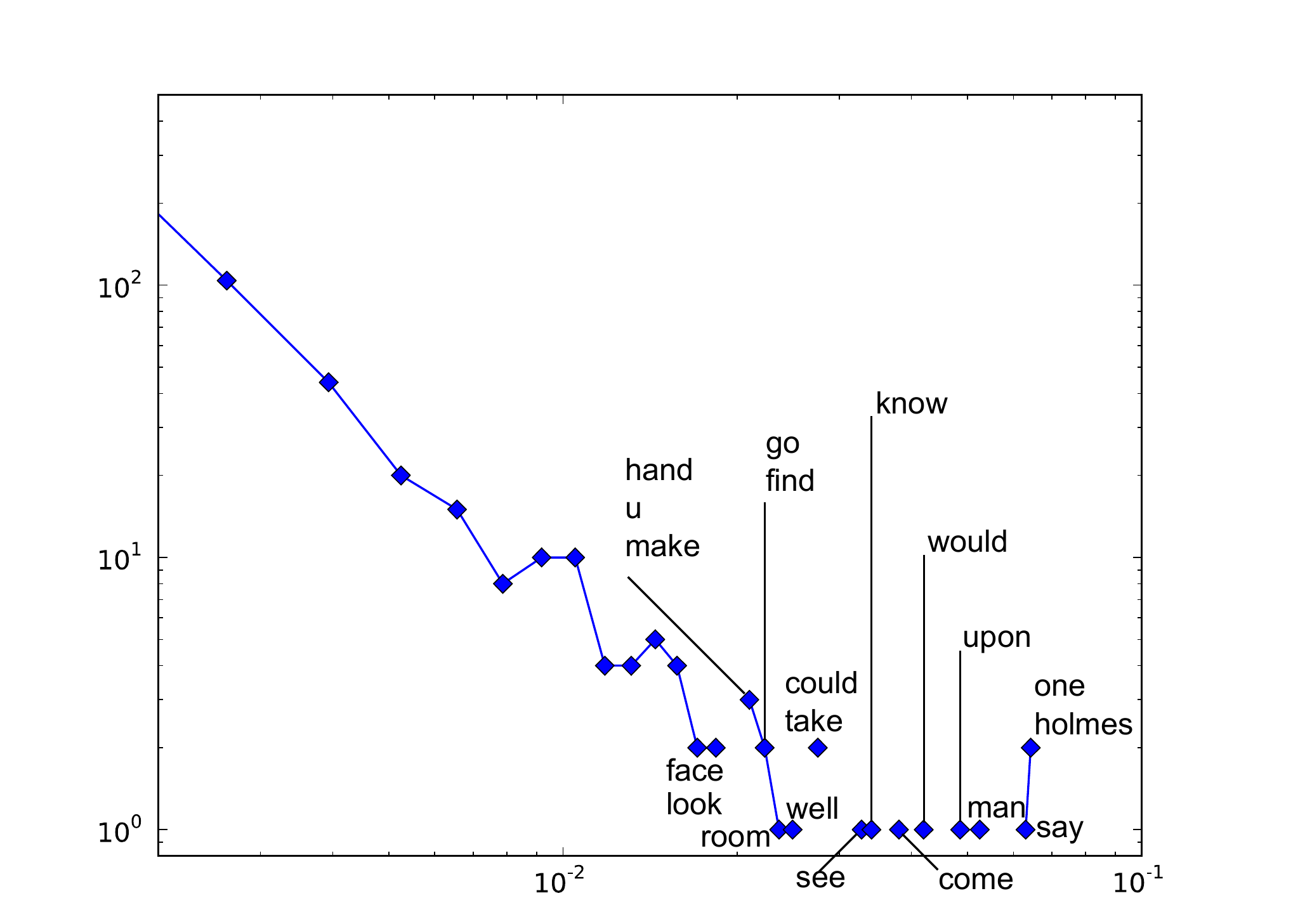}
\caption{Betweenness centrality distribution for ``The Return of Sherlock Holmes''. The top 20 nodes are labeled with the corresponding words.}
\label{fig:bc_distibution_2}
\end{figure}

Figures \ref{fig:bc_distibution_1} and \ref{fig:bc_distibution_2} show the distributions of betweenness centralities for two books from Arthur Conan Doyle: The Memoirs of Sherlock Holmes and The Return of Sherlock Holmes, which could be expected to be similar since these books were written by the same author in the same series of novels. In both figures, the highest centralities belong to the same words (19 out of 20) which also occupy almost the same relative positions. We shall therefore test the hypothesis that not only the frequency of usage but also the long-scale topological metrics of the organization of words may be reliable signatures of authorship.

To quantify this we introduce a similarity measure between pairs of texts as follows: for each network metric considered we give a rank $R$ to each word $w$ from a subset $V$ of top words with unique properties. In our approach the importance of a word depends on the particular network metric considered. As mentioned above, we select the words with the highest connectivity and betweenness centrality, and with the smallest values of shortest path and intermittency. For these two latter metrics the smallest values were chosen because their distributions present power laws with positive coefficients. We choose sets of $100$ top words since in subsidiary experiments we observed that the interval between $50-150$ words gave the best results. A ranking is assigned to each word, starting with the maximum value ($100$ in this case) for the word with the most extreme value (e.g. ``say'' for the betweenness centrality of ``The Memoirs of Sherlock Holmes''), and decreasing in one unit for each consecutive word until reaching the last of the top words which receives a ranking value of one. With these rankings, the similarity between two texts $A$ and $B$ for a given network metric is given by
\begin{equation}
A \cdot B = \sum_{w\in V_A \cap V_B}{R_A(w)R_B(w)},
\label{eq:similarity_measure}
\end{equation}
that is, if a word is present in the top words subsets of both texts, the product of its rankings adds to their similarity.

This similarity metric is guaranteed to be high only if the same words occupy similar positions in the distributions such as in figures \ref{fig:bc_distibution_1} and \ref{fig:bc_distibution_2}, with higher influence from the highest-ranked words. Equation \ref{eq:similarity_measure} implies that the norm or similarity of a text with itself is always the same, that is, $A\cdot A = B\cdot B = n(n+1)(2n+1)/6$, where $n$ is the size of $V$. We therefore normalize all similarities for this value to be one and the minimum value to be zero, and define the distance $D_{AB}$ between two texts as being one minus this normalized value. It is worth noting that other similarity metrics could be used to compare two pairs of texts, but the dot product adopted here appears to be the most straightforward, as it is done in bag-of-words methods~\cite{manning2008introduction}.

\begin{figure}
\begin{center}
\includegraphics[width=0.75\columnwidth]{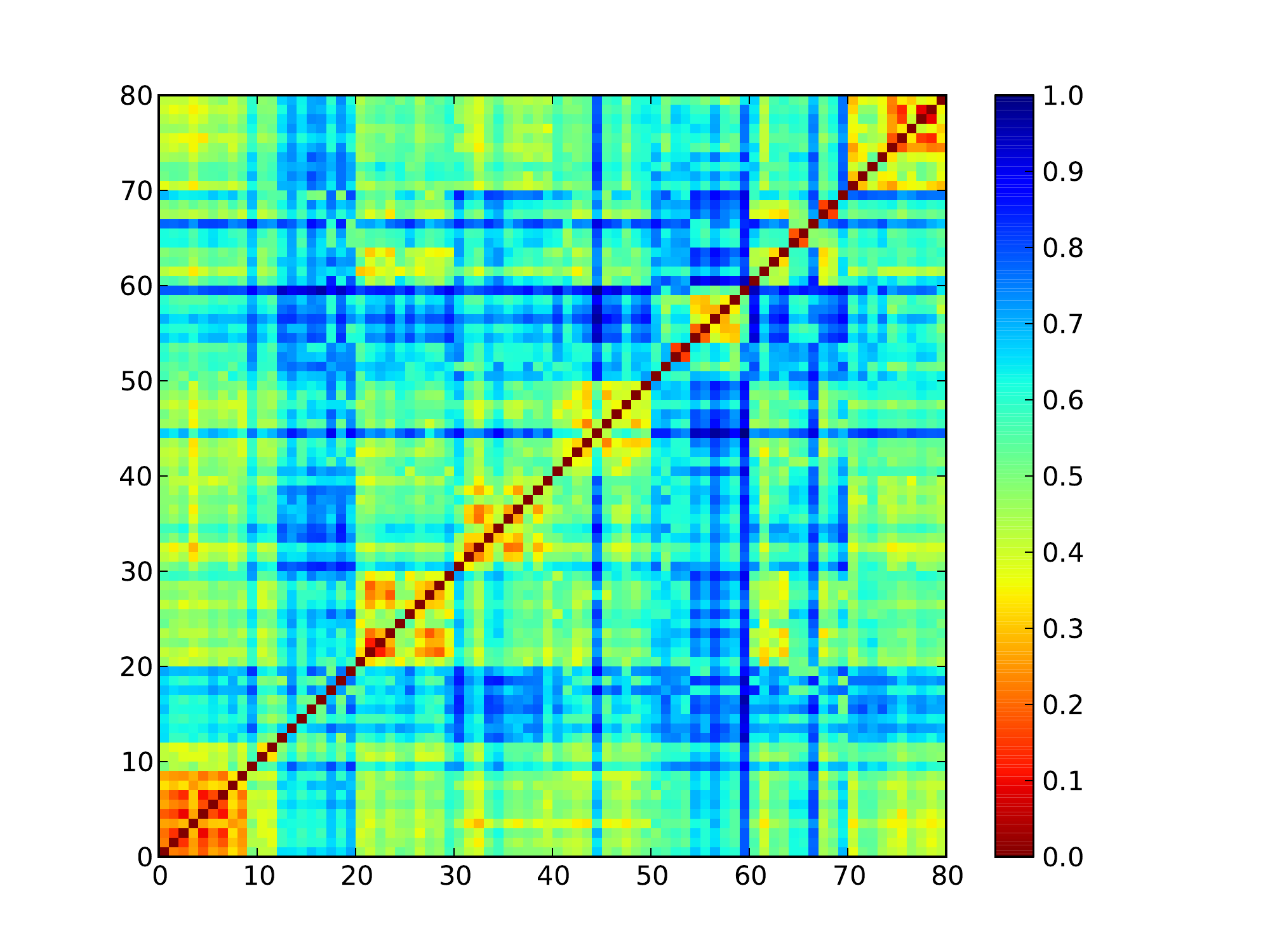}
\end{center}
\caption{Betweenness centrality distance matrix for the second collection, each decade corresponds to texts from the same author.}
\label{fig:distance_matrix}
\end{figure}

With all the values $D_{AB}$ we produce a distance matrix for each metric. The distance matrix for the betweenness centrality of one of the collections is shown in figure \ref{fig:distance_matrix}, where the indices $0$ to $9$ correspond to texts from the first author, texts $10$ to $19$ to the second author and so on. Note that, in general, texts from the same author appear to be closer among themselves compared to texts from different authors even if they are relatively separated (e.g. texts $10-19$ and $50-59$).

\subsection{Combining Distance Matrices}

One strength of the approach is the ability of observing different aspects of the network structure simultaneously. Each metric yields a different distance matrix; hence, we can observe the similarity between texts at different scales. We now combine information from the distinct metrics in order to have useful data for the classification algorithms.

\begin{figure}
%fig_dm_100_bc_jet_r.png
\centering
\includegraphics[width=0.75\columnwidth]{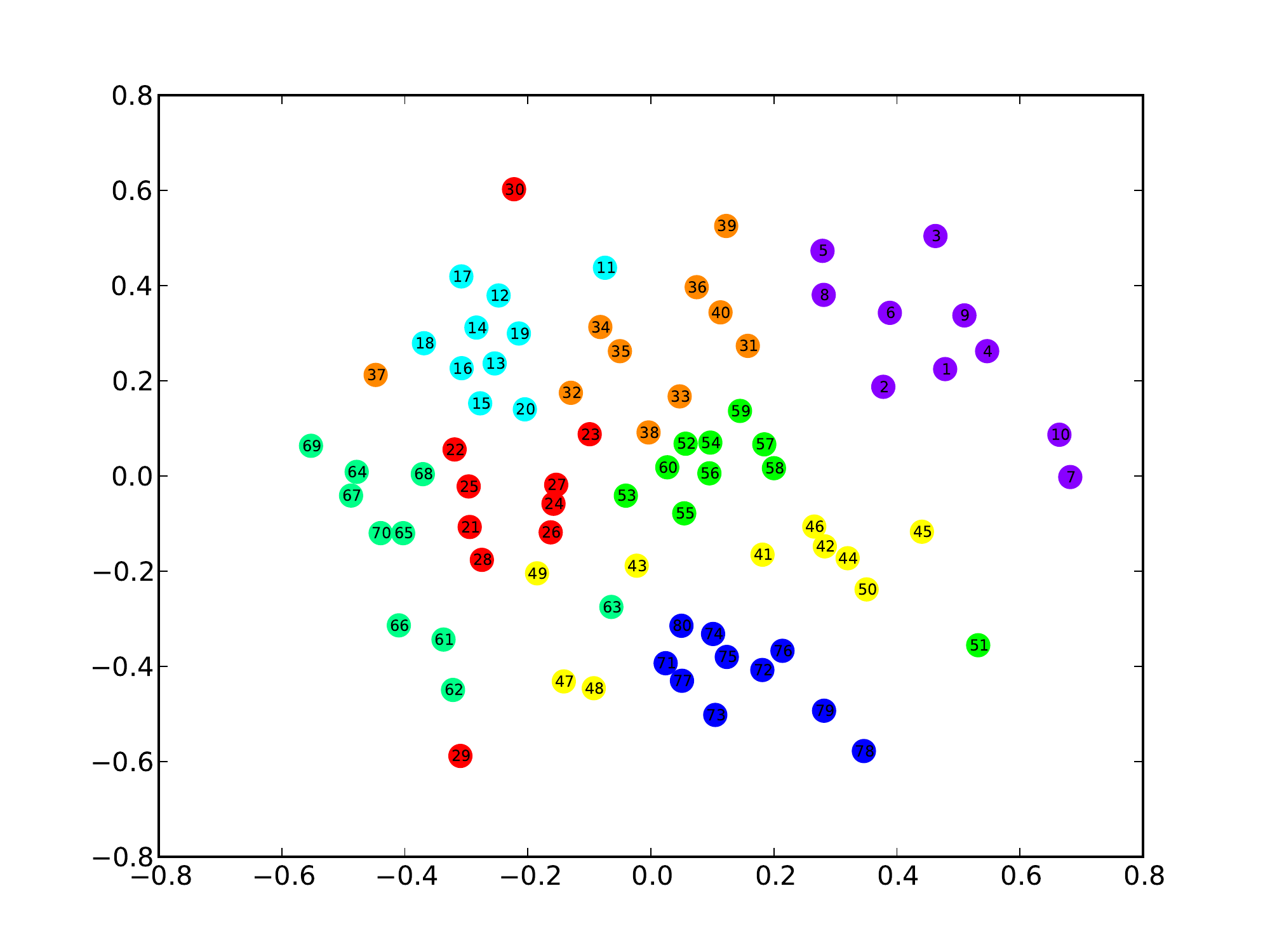}
\caption{Bi-dimensional MDS mapping of the betweenness centrality distances for the third collection. Numbers correspond to texts indices, colors correspond to authors.}
\label{fig:mds}
\end{figure}

In this study we employed two strategies for the input into the classification algorithms. In the first, we simply used the whole of the distance matrices for the different metrics, i.e. with distances as attributes. In the second strategy, we reduced the dimensionality of the distance metrics with Multi-dimensional scaling (MDS) \cite{kruskal1964multidimensional}, with the aim of capturing the highest similarities while eliminating possible unnecessary information that may harm the classification task. MDS was conceived to map distances into positions in a space so that the distances between these positions reproduce as well as possible the original input distances. The space obtained is usually intended to have a small dimensionality and the algorithm is largely used for visualization purposes. The positions obtained when applying the algorithm to map one of the distance matrices to a two-dimensional space is presented in figure \ref{fig:mds} reflecting the similarities between same-author texts already observed with the distance matrices. We use MDS to map the four distance matrices of each collection into four subspaces and then join these subspaces into a space of bigger dimension: if we write the positions in each subspace as a matrix $M\times N_i$ where $M$ is the number of points ($80$ texts per collection in our case) and $N_i$ is the dimensionality of the subspace, then the positions in the total space are given by a matrix $M\times (N_1+N_2+N_3+N_4)$ where each row is composed joining head to tail the corresponding rows of the positions on the subspaces.

Instead of a bi-dimensional mapping such as that of figure \ref{fig:mds}, the dimensionalities $N_i$ are calculated based on the stress or cost function (the difference between the actual and the obtained distances). As stress is a monotonically decreasing function of the number of dimensions we set a threshold of $10\%$ of the value for one dimension (also known as the elbow method) which was usually found to be reached at $N_i=6$.

\subsection{Data analysis}

The final positions on the composed space are the attributes for the data analysis algorithms. Analysis is done with supervised learning algorithms from the main types currently in use: tree-based J48; K-Nearest Neighbors (KNN); Naive Bayesian (NB); and Radial Basis Function Network (RBFN). For all cases $10$-fold cross validation is applied and the parameters are set to their default values~\cite{amancio2014systematic}. For KNN the number of neighbors is set to three which is the smallest odd non-trivial value. For RBFN the number of clusters is set to eight which is the number of authors. Authorship is also addressed using the standard TF-IDF model. Since TF-IDF returns a distance matrix, we also use MDS in this single matrix in order to apply the same classification algorithms on both approaches.

\section{Results and Discussion}

\begin{figure}
\centering
\includegraphics[width=0.75\columnwidth]{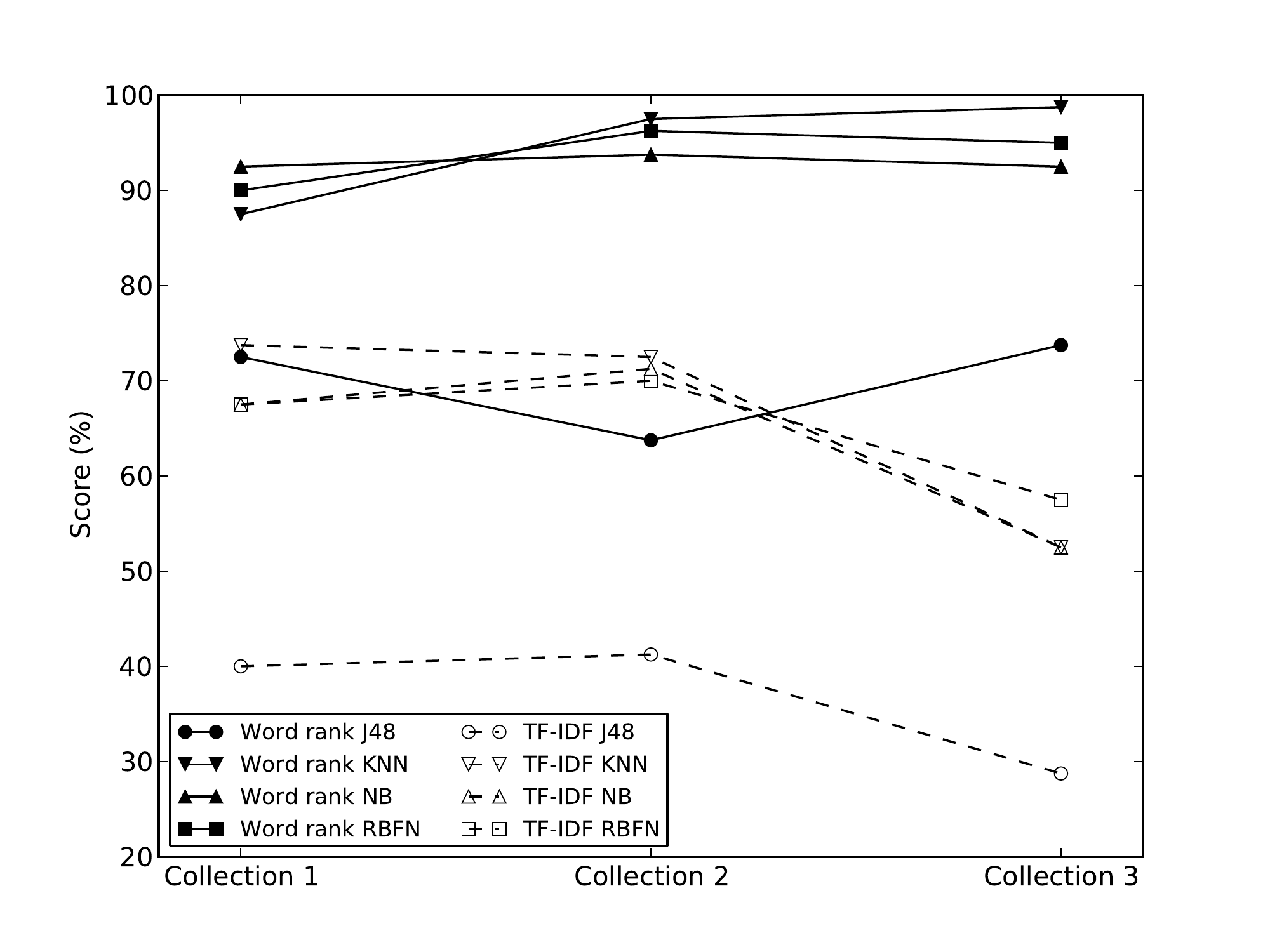}
\caption{Scores obtained with the method introduced here, also using MDS, and with TF-IDF for the three collections of texts.}
\label{fig:scores}
\end{figure}

The approach based on distance matrices as input for the classification algorithms was applied for the three collections, for which the success score for classification by chance is $1/8=12.5\%$. The results are outstanding as shown in table \ref{table:scores}, especially when MDS was used. It seems therefore that reducing the dimensionality actually amounted to an efficient feature selection, probably eliminating data that brought noise to the analysis. With MDS, typical accuracy rates were above $90\%$ and the maximum value was $98.75\%$ obtained with KNN for the third collection which corresponds to only one text (out of $80$) not correctly classified. These scores greatly surpassed the values obtained by applying the TF-IDF method, for which the mean scores among collections were $36.67\%$ for J48, $66.25\%$ for KNN, $63.75\%$ for NB, and $65\%$ for RBFN, as shown in figure \ref{fig:scores}. These scores demonstrate the added value of using the network structure over relying only on the frequency of appearance of features. Significantly, the higher scores for the approach introduced here are maintained when changing the classification algorithm (KNN, NB, and RBFN), which indicates the robustness of the proposed metrics. Also worth noting is that the present approach outperforms a previous one where the topology of networks was taken without considering the labels of the nodes (words) \cite{akimushkin2017plos}, for which the accuracy rates for the second collection studied here were $63.75\%$ with J48, $88.75\%$ with KNN, $81.25\%$ with NB, and $83.75\%$ with RBFN. In addition, the approach presented is less demanding, both computationally and conceptually, than the previous one.

\begin{table}%
\centering
\begin{tabular}{lcccc}
& J48 & KNN & NB & RBFN \\
\hline
Without MDS & ~ & ~ & ~ & ~ \\
Collection 1 & 73.75 & 85.00 & 85.00 & 62.50 \\
Collection 2 & 73.75 & 83.75 & 82.50 & 78.75 \\
Collection 3 & 75.00 & 92.50 & 80.00 & 71.25 \\
\hline
Using MDS & ~ & ~ & ~ & ~ \\
Collection 1 & 72.50 & 87.50 & 92.50 & 90.00 \\
Collection 2 & 63.75 & 97.50 & 93.75 & 96.25 \\
Collection 3 & 73.75 & 98.75 & 92.50 & 95.00 \\
\hline
\end{tabular}
\caption{Accuracy rates (in percentage) in identifying the authors in the three collections, using several machine learning algorithms. Results are shown with the input comprising the whole distance matrices (without MDS) and applying MDS on the matrices. For the three last algorithms MDS improved accuracy in all cases.}
\label{table:scores}
\end{table}

Taken together, the results indicate that, apart from the frequency of appearance and syntactical relations, certain words are essential to the structure of a text as a whole. The procedure to identify such words using complex networks has been successful, since utilization of these words is author-dependent. The co-occurrence network procedure allows one to observe the features of a word at different scales in the text. For instance, words with low intermittency, i.e. whose appearance in the text is highly periodic, had a high relevance for the authorship attribution, even though a word can have a low intermittency and not appear in most paragraphs.

Ours is a unified framework for multivariate analysis of texts. In contrast to other multivariate approaches, the generalized similarity measure \ref{eq:similarity_measure} allows to easily introduce new features to the scheme using some of the many node-local network metrics in existence. Care must be taken, however, because not all metrics are useful and some can even lower the performance. For example, we tested the clustering coefficient (characterized by a bell-shaped distribution) and eigenvalue centrality without success. Even though the computation of the similarity measures between documents resembles that of TF-IDF (i.e. cosine similarity) there are significant differences, mainly the fact that the vectors are of a much smaller size and that there are no repeated values. A question to be further studied is the optimal ranking procedure: we chose a ranking \textit{a la} Zipf because of the presence of power law distributions, but other rankings could be possible. While both TF-IDF and our method account for the heterogeneity of sizes of texts, our ranking procedure has two principal advantages: computation is faster and most importantly, the ranking does not have to be repeated every time the collection is modified, which is especially advantageous with big collections.

\section{Conclusions}

We have introduced an approach by which the representation of text with complex networks is enhanced by considering the words corresponding to the nodes. This is done with a similarity metric to compare two pieces of text where the presence of the most relevant words, according to network metrics, is taken into account. When the distance matrices obtained with the similarity metrics were used as input into machine learning algorithms, a high accuracy was achieved which reached $98.75\%$ for one of the book collections. Significantly, the accuracy was considerably higher than for traditional methods based on TF-IDF, being also higher than other network approaches that did not consider the label of the nodes. Also relevant is that the performance was improved with dimensionality reduction with MDS, which is advantageous owing to the lower computational cost.

With regard to the limitations, one should emphasize that the present approach is not useful for very short texts (such as a summary of an article). The method can be extended to employ other metrics and multi-node distributions. As some authors have pointed out \cite{van2005new}, it is likely that every person has a characteristic writing fingerprint owing to their particular way to learn a language. If this is the case, the traits that define such fingerprint are probably complex and not bounded to one single measure. Finally, the approach proposed could be used for such other applications as part-of-speech analysis of network distributions and resolution of word polysemy.

\section*{Acknowledgments}

This work was supported by CNPq (Brazil) and FAPESP (grants 2014/20830-0, 2013/14262-7 and 2016/19069-9).

%\bibliography{bibliography}

\section*{References}

\begin{thebibliography}{36}
\expandafter\ifx\csname natexlab\endcsname\relax\def\natexlab#1{#1}\fi
\providecommand{\url}[1]{\texttt{#1}}
\providecommand{\href}[2]{#2}
\providecommand{\path}[1]{#1}
\providecommand{\DOIprefix}{doi:}
\providecommand{\ArXivprefix}{arXiv:}
\providecommand{\URLprefix}{URL: }
\providecommand{\Pubmedprefix}{pmid:}
\providecommand{\doi}[1]{\href{http://dx.doi.org/#1}{\path{#1}}}
\providecommand{\Pubmed}[1]{\href{pmid:#1}{\path{#1}}}
\providecommand{\bibinfo}[2]{#2}
\ifx\xfnm\relax \def\xfnm[#1]{\unskip,\space#1}\fi
%Type = Article
\bibitem[{Liang(2017)}]{Liang2017802}
\bibinfo{author}{W.~Liang}, \bibinfo{journal}{Physica A} \bibinfo{volume}{468}
  (\bibinfo{year}{2017}) \bibinfo{pages}{802 -- 808}.
%Type = Article
\bibitem[{Silva and Amancio(2012)}]{0295-5075-98-5-58001}
\bibinfo{author}{T.~C. Silva}, \bibinfo{author}{D.~R. Amancio},
  \bibinfo{journal}{EPL (Europhysics Letters)} \bibinfo{volume}{98}
  (\bibinfo{year}{2012}) \bibinfo{pages}{58001}.
%Type = Article
\bibitem[{Zhong et~al.(2017)Zhong, Liu, Gao, and Wu}]{Zhong2017462}
\bibinfo{author}{X.~Zhong}, \bibinfo{author}{J.~Liu}, \bibinfo{author}{Y.~Gao},
  \bibinfo{author}{L.~Wu}, \bibinfo{journal}{Physica A} \bibinfo{volume}{466}
  (\bibinfo{year}{2017}) \bibinfo{pages}{462 -- 475}.
%Type = Book
\bibitem[{Manning et~al.(2008)Manning, Raghavan, Sch{\"u}tze
  et~al.}]{manning2008introduction}
\bibinfo{author}{C.~D. Manning}, \bibinfo{author}{P.~Raghavan},
  \bibinfo{author}{H.~Sch{\"u}tze}, et~al., \bibinfo{title}{Introduction to
  information retrieval}, volume~\bibinfo{volume}{1},
  \bibinfo{publisher}{Cambridge university press Cambridge},
  \bibinfo{year}{2008}.
%Type = Article
\bibitem[{Asghar et~al.(2017)Asghar, Khan, Ahmad, Qasim, and
  Khan}]{10.1371/journal.pone.0171649}
\bibinfo{author}{M.~Z. Asghar}, \bibinfo{author}{A.~Khan},
  \bibinfo{author}{S.~Ahmad}, \bibinfo{author}{M.~Qasim},
  \bibinfo{author}{I.~A. Khan}, \bibinfo{journal}{PLoS One}
  \bibinfo{volume}{12} (\bibinfo{year}{2017}) \bibinfo{pages}{e0171649}.
%Type = Article
\bibitem[{Amancio et~al.(2012)Amancio, Oliveira~Jr., and Costa}]{Inform2012427}
\bibinfo{author}{D.~R. Amancio}, \bibinfo{author}{O.~N. Oliveira~Jr.},
  \bibinfo{author}{L.~F. Costa}, \bibinfo{journal}{Journal of Informetrics}
  \bibinfo{volume}{6} (\bibinfo{year}{2012}) \bibinfo{pages}{427 -- 434}.
%Type = Article
\bibitem[{Viana et~al.(2013)Viana, Amancio, and Costa}]{Viana2013371}
\bibinfo{author}{M.~P. Viana}, \bibinfo{author}{D.~R. Amancio},
  \bibinfo{author}{L.~F. Costa}, \bibinfo{journal}{Journal of Informetrics}
  \bibinfo{volume}{7} (\bibinfo{year}{2013}) \bibinfo{pages}{371 -- 378}.
%Type = Article
\bibitem[{Juola(2006)}]{juola2006authorship}
\bibinfo{author}{P.~Juola}, \bibinfo{journal}{Foundations and Trends in
  information Retrieval} \bibinfo{volume}{1} (\bibinfo{year}{2006})
  \bibinfo{pages}{233--334}.
%Type = Article
\bibitem[{Stamatatos(2009)}]{Stamatatos:2009:SMA:1527090.1527102}
\bibinfo{author}{E.~Stamatatos}, \bibinfo{journal}{J. Am. Soc. Inf. Sci.
  Technol.} \bibinfo{volume}{60} (\bibinfo{year}{2009})
  \bibinfo{pages}{538--556}.
%Type = Article
\bibitem[{Amancio et~al.(2011)Amancio, Altmann, Oliveira~Jr, and
  Costa}]{amancio2011comparing}
\bibinfo{author}{D.~R. Amancio}, \bibinfo{author}{E.~G. Altmann},
  \bibinfo{author}{O.~N. Oliveira~Jr}, \bibinfo{author}{L.~F. Costa},
  \bibinfo{journal}{New Journal of Physics} \bibinfo{volume}{13}
  (\bibinfo{year}{2011}) \bibinfo{pages}{123024}.
%Type = Article
\bibitem[{Amancio(2015)}]{1742-5468-2015-3-P03005}
\bibinfo{author}{D.~R. Amancio}, \bibinfo{journal}{Journal of Statistical
  Mechanics: Theory and Experiment} \bibinfo{volume}{2015}
  (\bibinfo{year}{2015}) \bibinfo{pages}{P03005}.
%Type = Inproceedings
\bibitem[{Peng et~al.(2003)Peng, Schuurmans, Wang, and
  Keselj}]{peng2003language}
\bibinfo{author}{F.~Peng}, \bibinfo{author}{D.~Schuurmans},
  \bibinfo{author}{S.~Wang}, \bibinfo{author}{V.~Keselj}, in:
  \bibinfo{booktitle}{Proceedings of the tenth conference on European chapter
  of the Association for Computational Linguistics-Volume 1},
  \bibinfo{organization}{Association for Computational Linguistics}, pp.
  \bibinfo{pages}{267--274}.
%Type = Inproceedings
\bibitem[{Escalante et~al.(2011)Escalante, Solorio, and Montes-y
  G{\'o}mez}]{escalante2011local}
\bibinfo{author}{H.~J. Escalante}, \bibinfo{author}{T.~Solorio},
  \bibinfo{author}{M.~Montes-y G{\'o}mez}, in: \bibinfo{booktitle}{Proceedings
  of the 49th Annual Meeting of the Association for Computational Linguistics:
  Human Language Technologies-Volume 1}, \bibinfo{organization}{Association for
  Computational Linguistics}, pp. \bibinfo{pages}{288--298}.
%Type = Inproceedings
\bibitem[{Forstall and Scheirer(2010)}]{forstall2010features}
\bibinfo{author}{C.~Forstall}, \bibinfo{author}{W.~Scheirer}, in:
  \bibinfo{booktitle}{Journal of the Chicago Colloquium on Digital Humanities
  and Computer Science}, volume~\bibinfo{volume}{1}.
%Type = Article
\bibitem[{Kukushkina et~al.(2001)Kukushkina, Polikarpov, and
  Khmelev}]{kukushkina2001using}
\bibinfo{author}{O.~V. Kukushkina}, \bibinfo{author}{A.~Polikarpov},
  \bibinfo{author}{D.~V. Khmelev}, \bibinfo{journal}{Problems of Information
  Transmission} \bibinfo{volume}{37} (\bibinfo{year}{2001})
  \bibinfo{pages}{172--184}.
%Type = Article
\bibitem[{Chaski(2005)}]{chaski2005s}
\bibinfo{author}{C.~E. Chaski}, \bibinfo{journal}{International journal of
  digital evidence} \bibinfo{volume}{4} (\bibinfo{year}{2005})
  \bibinfo{pages}{1--13}.
%Type = Article
\bibitem[{Harris(1954)}]{Harris:54}
\bibinfo{author}{Z.~Harris}, \bibinfo{journal}{Word} \bibinfo{volume}{10}
  (\bibinfo{year}{1954}) \bibinfo{pages}{146--62}.
%Type = Book
\bibitem[{Zipf(1935)}]{zipf1935psycho}
\bibinfo{author}{G.~K. Zipf}, \bibinfo{title}{The psycho-biology of language},
  \bibinfo{publisher}{Houghton, Mifflin}, \bibinfo{year}{1935}.
%Type = Article
\bibitem[{Ferrer-i Cancho and Sol{\'e}(2001)}]{ferrer2001two}
\bibinfo{author}{R.~Ferrer-i Cancho}, \bibinfo{author}{R.~V. Sol{\'e}},
  \bibinfo{journal}{Journal of Quantitative Linguistics} \bibinfo{volume}{8}
  (\bibinfo{year}{2001}) \bibinfo{pages}{165--173}.
%Type = Article
\bibitem[{Amancio et~al.(2014)Amancio, Comin, Casanova, Travieso, Bruno,
  Rodrigues, and Costa}]{amancio2014systematic}
\bibinfo{author}{D.~R. Amancio}, \bibinfo{author}{C.~H. Comin},
  \bibinfo{author}{D.~Casanova}, \bibinfo{author}{G.~Travieso},
  \bibinfo{author}{O.~M. Bruno}, \bibinfo{author}{F.~A. Rodrigues},
  \bibinfo{author}{L.~F. Costa}, \bibinfo{journal}{PLoS One}
  \bibinfo{volume}{9} (\bibinfo{year}{2014}) \bibinfo{pages}{e94137}.
%Type = Article
\bibitem[{Sparck~Jones(1972)}]{sparck1972statistical}
\bibinfo{author}{K.~Sparck~Jones}, \bibinfo{journal}{Journal of documentation}
  \bibinfo{volume}{28} (\bibinfo{year}{1972}) \bibinfo{pages}{11--21}.
%Type = Article
\bibitem[{Zhang et~al.(2007)Zhang, Marsza{\l}ek, Lazebnik, and
  Schmid}]{zhang2007local}
\bibinfo{author}{J.~Zhang}, \bibinfo{author}{M.~Marsza{\l}ek},
  \bibinfo{author}{S.~Lazebnik}, \bibinfo{author}{C.~Schmid},
  \bibinfo{journal}{International journal of computer vision}
  \bibinfo{volume}{73} (\bibinfo{year}{2007}) \bibinfo{pages}{213--238}.
%Type = Inproceedings
\bibitem[{Ke{\v{s}}elj et~al.(2003)Ke{\v{s}}elj, Peng, Cercone, and
  Thomas}]{kevselj2003n}
\bibinfo{author}{V.~Ke{\v{s}}elj}, \bibinfo{author}{F.~Peng},
  \bibinfo{author}{N.~Cercone}, \bibinfo{author}{C.~Thomas}, in:
  \bibinfo{booktitle}{Proceedings of the conference pacific association for
  computational linguistics, PACLING}, volume~\bibinfo{volume}{3}, pp.
  \bibinfo{pages}{255--264}.
%Type = Article
\bibitem[{Clement and Sharp(2003)}]{clement2003ngram}
\bibinfo{author}{R.~Clement}, \bibinfo{author}{D.~Sharp},
  \bibinfo{journal}{Literary and linguistic computing} \bibinfo{volume}{18}
  (\bibinfo{year}{2003}) \bibinfo{pages}{423--447}.
%Type = Article
\bibitem[{Baayen et~al.(1996)Baayen, Van~Halteren, and
  Tweedie}]{baayen1996outside}
\bibinfo{author}{H.~Baayen}, \bibinfo{author}{H.~Van~Halteren},
  \bibinfo{author}{F.~Tweedie}, \bibinfo{journal}{Literary and Linguistic
  Computing} \bibinfo{volume}{11} (\bibinfo{year}{1996})
  \bibinfo{pages}{121--132}.
%Type = Inproceedings
\bibitem[{Rygl et~al.(2012)Rygl, Zemkov{\'a}, and
  Kov{\'a}r}]{rygl2012authorship}
\bibinfo{author}{J.~Rygl}, \bibinfo{author}{K.~Zemkov{\'a}},
  \bibinfo{author}{V.~Kov{\'a}r}, in: \bibinfo{booktitle}{Proceedings of Sixth
  Workshop on Recent Advances in Slavonic Natural Language Processing, RASLAN},
  pp. \bibinfo{pages}{111--119}.
%Type = Article
\bibitem[{Cong and Liu(2014)}]{Cong2014598}
\bibinfo{author}{J.~Cong}, \bibinfo{author}{H.~Liu}, \bibinfo{journal}{Physics
  of Life Reviews} \bibinfo{volume}{11} (\bibinfo{year}{2014})
  \bibinfo{pages}{598 -- 618}.
%Type = Article
\bibitem[{Dorogovtsev and Mendes(2001)}]{dorogovtsev2001language}
\bibinfo{author}{S.~N. Dorogovtsev}, \bibinfo{author}{J.~F.~F. Mendes},
  \bibinfo{journal}{Proceedings of the Royal Society of London. Series B:
  Biological Sciences} \bibinfo{volume}{268} (\bibinfo{year}{2001})
  \bibinfo{pages}{2603--2606}.
%Type = Inproceedings
\bibitem[{Choudhury et~al.(2010)Choudhury, Chatterjee, and
  Mukherjee}]{choudhury2010global}
\bibinfo{author}{M.~Choudhury}, \bibinfo{author}{D.~Chatterjee},
  \bibinfo{author}{A.~Mukherjee}, in: \bibinfo{booktitle}{Proceedings of the
  23rd International Conference on Computational Linguistics: Posters},
  \bibinfo{organization}{Association for Computational Linguistics}, pp.
  \bibinfo{pages}{162--170}.
%Type = Article
\bibitem[{Mehri et~al.(2012)Mehri, Darooneh, and Shariati}]{mehri2012complex}
\bibinfo{author}{A.~Mehri}, \bibinfo{author}{A.~H. Darooneh},
  \bibinfo{author}{A.~Shariati}, \bibinfo{journal}{Physica A}
  \bibinfo{volume}{391} (\bibinfo{year}{2012}) \bibinfo{pages}{2429--2437}.
%Type = Article
\bibitem[{Akimushkin et~al.(2017)Akimushkin, Amancio, and
  Oliveira~Jr.}]{akimushkin2017plos}
\bibinfo{author}{C.~Akimushkin}, \bibinfo{author}{D.~R. Amancio},
  \bibinfo{author}{O.~N. Oliveira~Jr.}, \bibinfo{journal}{PLoS One}
  \bibinfo{volume}{12} (\bibinfo{year}{2017}) \bibinfo{pages}{e0170527}.
%Type = Book
\bibitem[{Borg and Groenen(2005)}]{BorgGroenen2005}
\bibinfo{author}{I.~Borg}, \bibinfo{author}{P.~Groenen},
  \bibinfo{title}{{Modern Multidimensional Scaling: Theory and Applications}},
  \bibinfo{publisher}{Springer}, \bibinfo{year}{2005}.
%Type = Unpublished
\bibitem[{Greene and Rubin(1971)}]{GreRub}
\bibinfo{author}{B.~B. Greene}, \bibinfo{author}{G.~M. Rubin},
  \bibinfo{title}{Automatic grammatical tagging of english},
  \bibinfo{year}{1971}. \bibinfo{note}{Department of Linguistics, Brown
  University, Providence, Rhode Island}.
%Type = Article
\bibitem[{Ortuno et~al.(2002)Ortuno, Carpena, Bernaola-Galván, Muñoz, and
  Somoza}]{0295-5075-57-5-759}
\bibinfo{author}{M.~Ortuno}, \bibinfo{author}{P.~Carpena},
  \bibinfo{author}{P.~Bernaola-Galván}, \bibinfo{author}{E.~Muñoz},
  \bibinfo{author}{A.~M. Somoza}, \bibinfo{journal}{EPL (Europhysics Letters)}
  \bibinfo{volume}{57} (\bibinfo{year}{2002}) \bibinfo{pages}{759}.
%Type = Article
\bibitem[{Kruskal(1964)}]{kruskal1964multidimensional}
\bibinfo{author}{J.~B. Kruskal}, \bibinfo{journal}{Psychometrika}
  \bibinfo{volume}{29} (\bibinfo{year}{1964}) \bibinfo{pages}{1--27}.
%Type = Article
\bibitem[{Van~Halteren et~al.(2005)Van~Halteren, Baayen, Tweedie, Haverkort,
  and Neijt}]{van2005new}
\bibinfo{author}{H.~Van~Halteren}, \bibinfo{author}{H.~Baayen},
  \bibinfo{author}{F.~Tweedie}, \bibinfo{author}{M.~Haverkort},
  \bibinfo{author}{A.~Neijt}, \bibinfo{journal}{Journal of Quantitative
  Linguistics} \bibinfo{volume}{12} (\bibinfo{year}{2005})
  \bibinfo{pages}{65--77}.

\end{thebibliography}

\end{document}